\documentclass[letterpaper, 10 pt, conference]{ieeeconf}

\IEEEoverridecommandlockouts
\usepackage{tabularx}
\usepackage{graphicx}
\usepackage{amsmath}
\usepackage{array}
\usepackage[hidelinks]{hyperref}
\usepackage{ragged2e}
\usepackage{makecell}
\usepackage{multirow}
\usepackage{caption}
\title{\LARGE \bf
VTD: Visual and Tactile Database for Driver State and Behavior Perception
}

\author{Jie Wang$^{1}$,Mobing Cai$^{2}$,Zhongpan Zhu$^{*}$,\textit{Member,IEEE,}Hongjun Ding,Jiwei Yi and Aimin Du
\thanks{*This work was co-supported by the National Key Research and Development Pro-gram of China (No. 2021YFE0193900 and 2020AAA0108901), the National Natural Science Foundation of China (No. 52002286, 51975415 and 62088101), the Fundamental Research Funds for the Central Universities (22120220642). (Corresponding author: Zhongpan Zhu).}
\thanks{$^{1}$Jie Wang, Zhongpan Zhu are with the College of Electronics and Information Engineering, Tongji University, Shanghai 201804, China, and with Frontiers Science Center for Intelligent Autonomous Systems, Shanghai 201210,China(e-mail:2054310@tongji.edu.cn;521bergsteiger@tongji.edu.cn;)}%
\thanks{$^{2}$Mobing Cai is with the Department of Engineering and Trinity College, University of Oxford, Oxford, OX1 3BH, United Kingdom. (email:mobing.cai@trinity.ox.ac.uk)
       }%
\thanks{$^{}$Hongjun Ding is with CATARC Automotive Technology (Shanghai) Co., LTD.(email:dinghongjun@catarc.ac.cn), Jiwei Yi and Aimin Du are with the school of Automotive Studies, Tongji University, Shanghai 201804, China (e-mail:2033522@tongji.edu.cn;  duaimin@tongji.edu.cn;)
       }%
}

\begin{document}

\maketitle
\thispagestyle{empty}
\pagestyle{empty}

\begin{abstract}


In the domain of autonomous vehicles, the human-vehicle co-pilot system has garnered significant research attention. To address the subjective uncertainties in driver state and interaction behaviors, which are pivotal to the safety of Human-in-the-loop co-driving systems, we introduce a novel visual-tactile perception method. Utilizing a driving simulation platform, a comprehensive dataset has been developed that encompasses multi-modal data under fatigue and distraction conditions. The experimental setup integrates driving simulation with signal acquisition, yielding 600 minutes of fatigue detection data from 15 subjects and 102 takeover experiments with 17 drivers. The dataset, synchronized across modalities, serves as a robust resource for advancing cross-modal driver behavior perception algorithms.

\end{abstract}




%

\section{INTRODUCTION}

In recent years, data-driven autonomous vehicles have encountered SOTIF (Safety of the Intended Functionality) issues and long-tail challenges in their AI algorithms. These challenges arise due to the complexity and unpredictability of real-world scenarios where autonomous vehicles must navigate, requiring robust algorithms for safe operation under various conditions. Furthermore, the uncertainty in driver non-driving behavior\footnote{The uncertainty in driver non-driving behavior refers to situations where the driver is not fully engaged in the primary driving task due to factors like fatigue or distraction.} and abnormal state\footnote{Abnormal state refers to the unpredictability in driving behavior caused by changes in the driver's state, which elevates the risk in human-machine interaction.} within the context of human-vehicle co-driving further exacerbates the challenges faced by autonomous vehicles. Perceiving and predicting the driver’s uncertain behaviors accurately is crucial for developing effective AI algorithms that can adapt and respond appropriately.


As an individual with fully independent behavior possessing subjective initiative uncertainty and individual discrepancy, drivers may become increasingly reliant on autonomous systems as driving intelligence advances, leading to drowsiness, slower reaction times, and reduced risk perception\cite{li2022simulation,wu2023toward}. Furthermore, the addition of in-vehicle entertainment features introduces a potential risk of distracted driving. Research indicates that driver behavior is a key factor in most road traffic accidents, with fatigue and distraction being the primary causes\cite{arakawa2018trial,singh2021analyzing}. Changes in a driver's state increase the uncertainty of driving behavior and raise the risks associated with human-machine interaction, impeding the development of intelligent human-machine hybrid driving modes. Therefore, perceiving and understanding driver behavior and state has become a critical area for research breakthroughs.

To address these challenges, it is essential to collect high-quality datasets on fatigue and distraction, enhancing the ability to detect driver risks in intelligent co-driving systems and reducing potential dangers. These datasets should cover a wide range of driving conditions, human behavioral characteristics, and the specific interaction patterns found in human-machine co-driving scenarios.


The availability of human-vehicle co-driving datasets is crucial for the development and evaluation of artificial intelligence algorithms in autonomous vehicles. However, existing datasets for human-vehicle joint driving applications are still quite limited, especially in monitoring key states such as driver fatigue and distraction, and existing datasets are often limited in scope and depth, lacking the nuanced information required to address these challenges effectively. This paper proposes a multimodal cross-sensing method combining visual and haptic channels under controlled environmental conditions and constructs the VTD (Visual and Tactile Database for Driver State and Behavior Perception), a comprehensive, well-structured, large-scale dataset. The VTD dataset not only covers diverse driving conditions and human behavior characteristics but also places particular emphasis on monitoring and recording driver fatigue and distraction states. These data will help enhance the perception and understanding capabilities of autonomous driving systems regarding driver states in intelligent co-driving scenarios, thereby improving the safety and reliability of human-machine interaction and effectively reducing potential risks.

The contributions of VTD dataset are as follows:







1) This paper presents the VTD, a long-sequence multimodal natural driving dataset based on the fusion of visual and haptic data, which includes over 10 hours of fatigue driving data and 102 takeover scenarios. It effectively captures the multi-path driving conditions, as well as the mental and physical states and behavioral characteristics of drivers in multimodal environments.

2)The VTD is developed using a human-in-the-loop algorithm to ensure that the collected data accurately reflects real-world driving behavior. Additionally, the boundaries of driving environment are clearly defined, facilitating the quantification of the influencing mechanisms behind the driving behaviors of different groups.

3) Serving as a standardized platform for benchmarking in the field of human-vehicle co-driving, VTD offers valuable data support for cross-modal perception algorithms and scenarios related to driver fatigue and distraction. This capability significantly advances research in driver behavior perception.

\section{RELATED WORK}

In designing human-vehicle collaboration systems, recognizing the driver's intent, modeling behavior, and monitoring their state are critical. Despite advancements in autonomous driving technologies, the driver remains the core of system coordination\cite{gnatzig2012human}. Therefore, accurately monitoring the driver's fatigue and distraction is essential to ensure the safe operation of human-machine collaborative systems.

Driver state monitoring primarily focuses on physiological and psychological factors. Driver behavior monitoring involves analyzing specific behaviors during driving to infer their state and decision-making process. Behavior is an external manifestation of the state: the former reflects specific actions, while the latter represents the mental and physical condition. By observing driving behavior and measuring physiological and psychological indicators, a more comprehensive inference of the driver's fatigue or distraction can be made. This paper will introduce datasets in the field of human-vehicle collaboration from the perspectives of driver state and behavior, along with their applications and limitations in driver monitoring and human-vehicle co-driving system optimization.



\subsection{Datasets for Driver State}

As mentioned earlier, fatigue and distraction are the two main factors affecting driving safety. Therefore, this section will primarily focus on the current methods and datasets for monitoring driver distraction and fatigue, and analyze their applicability and limitations in real-world scenarios.


Research on driver distraction monitoring\footnote{NHTSA describes the distraction process as "any activity that diverts a driver’s attention away from the task of driving" and classifies it into visual, auditory, bio-mechanical, and cognitive distractions} has varied focuses. In recent years, the rise of AI algorithms for computer vision has led to a growing interest in analyzing driver behavior through visual perception methods. This includes studying facial expressions \cite{sigari2014review} and head gestures \cite{murphy2008head}. Key features commonly extracted in these studies are eye fixation duration, scan paths, eye-opening and eye-closing patterns, and head rotation angles. In addition, driving distraction monitoring and behavior analysis can also be achieved by using vehicle sensors to monitor vehicle conditions, such as steering wheel angle and accelerator pedal position, or by physiological indices such as driver  Electrocardiogram (ECG) \cite{deshmukh2017ecg,taherisadr2018ecg}, Electroencephalogram (EEG) \cite{li2021temporal,manikandan2021automobile,zuo2022driver}, Electromyography (EMG), and Galvanic Skin Response (GSR) \cite{manikandan2021automobile}.


Visual-based driver distraction monitoring accuracy is still limited in an actual driving environment due to problems including low resolution, motion blur, dynamic background, and occlusions \cite{jegham2020vision}. Hand movements, head gestures \cite{murphy2008head}, gaze direction \cite{jegham2018safe}, and pedal control are the keys to addressing the problems. Moreover, existing datasets still cannot characterize diverse, ambiguous, and personalized distraction behaviors influenced by the driver’s physiological and psychological state\cite{jegham2020novel,ortega2020dmd,das2015performance,schwarz2017driveahead,kopuklu2021driver,jegham2019mdad}. The investigations of the above datasets demonstrate a need for fine-grained distraction datasets with controllable and quantifiable conditions, multi-modal synchronized data, and data about driver distraction feature diversity.


Aside from driving distraction detection, driver fatigue is likewise an important research direction of driver behavior and state perception. Fatigue is displayed in various forms. Based on eye feature extractions, PERCLOS (percent eye closure), eye-white reflex, eye states, and yawning conditionsare included\cite{omidyeganeh2016yawning}. It can also be displayed through detection based on eye-mouth combinations \cite{ying2007monitoring}, eyelid closure and eye closure percentage combinations \cite{bergasa2006real}, FatigueTree \cite{yang2022fatigueview}, and other combinations.

We gathered and analyzed existing DMS datasets\cite{abtahi2014yawdd,pan2007eyeblink,weng2017driver,ghoddoosian2019realistic,weng2017driver,ortega2020dmd,diaz2016reduced,yang2022fatigueview} and discovered that in many datasets, fatigue is monitored by recording drivers’ facial features under natural driving conditions using RGB cameras. These natural-driving datasets face challenges in extracting driver fatigue features under nighttime dimmed lighting conditions or facial occlusions, and different types of physiological fatigue signals cannot be captured with a single vision. Differences in road environments may also lead to differences in driving loads, making it impossible to analyze the cause of fatigue and extract the differentiated impact on drivers. Additionally, most of the existing single-mode visual datasets about driver fatigue only concentrate on relatively monotonous visual features like blinking and yawning \cite{abtahi2014yawdd,pan2007eyeblink}, which lack time series and contextual features. Algorithms carried out on these single-mode datasets are too restrictive to be applied in reality and cannot contain all challenges\cite{yang2022fatigueview}. It is increasingly essential to determine how to provide greater flexibility and diversity in driver monitoring using multi-modal data to characterize fatigue signals and more complicated state combinations. However, the lack of a complete and comprehensive dataset in this field has bottlenecked the progress in algorithm development of driver fatigue detection \cite{ortega2020dmd}.


\subsection{Datasets for Driver Behaviors}

Driver behavior is distinct from the driver’s state. It refers to the specific movements a driver makes during the driving process, such as turning, accelerating, decelerating, and braking. Driver behavior can be captured and analyzed using vehicle sensors, cameras, and other monitoring devices. It is often associated with specific driving skills and traffic regulations, such as speeding, frequent lane changes, and running red lights. In essence, driver behavior encompasses the specific actions and maneuvers made by the driver, whereas driver state refers to their physical and mental condition. The driver’s state can be indirectly inferred by observing and evaluating their behaviors and by measuring relevant physiological indicators. Modeling and understanding a driver’s state through their behavior is essential for ensuring safety and facilitating assisted driving \cite{liu2021review}. To enhance driving safety, future intelligent vehicles should be capable of autonomously assessing the driver's behavior and competence using onboard sensors and operational data.


\begin{figure}[htbp]
  \centering
  \includegraphics[width=0.48\textwidth]{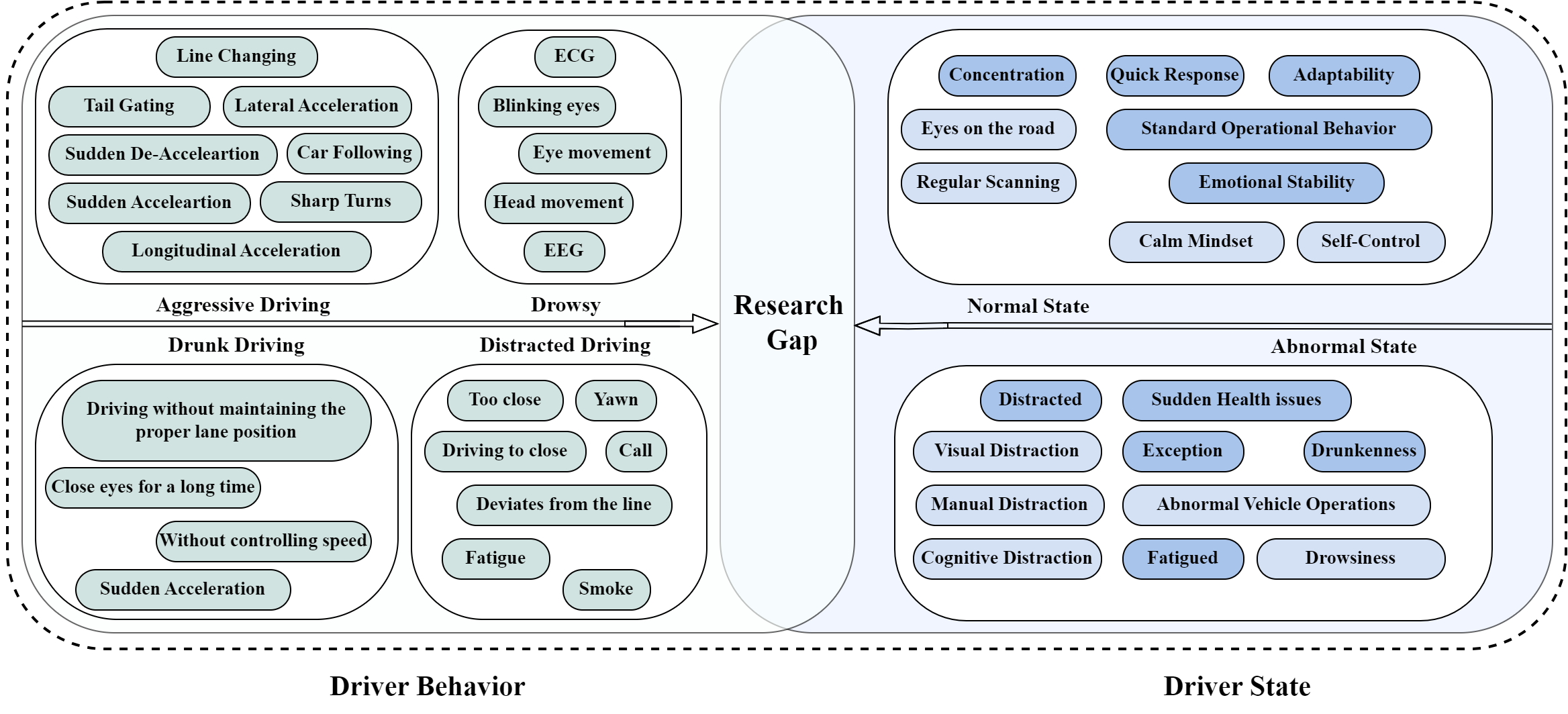}
  \caption{Research Gap in Driver Behavior and Driver State}
  \label{fig:example}
\end{figure}

Natural driving data is a vital resource for learning and understanding driving behaviors \cite{liu2021review}. Vehicle operations and driver behavior data are captured and collected through cameras and sensor arrays, providing essential support for research in this field. However, natural driving datasets often lack a unified boundary \cite{alkinani2020detecting,hu2010modeling}, and there is limited coupling of data on driver behaviors and states, making it challenging to perform quantitative analysis on contributing factors. Additionally, most datasets are outdated and suffer from low data quality due to limited device accuracy, rendering them inadequate for current needs. Publicly available natural driving datasets are also scarce because of the high cost of data collection\footnote{UTDrive, a large-scale natural dataset, includes driving data from 500 drivers across three countries, covering vehicles, sensors, and routes. In addition to UTDrive, the SHRP2 and MIT-AVT projects also focus on gathering natural driving data.}. The VTD dataset aims to provide new options for advancing research in this field.


\section{METHODOLOGY}

This section proposes a multi-view and multi-modal database named VTD for studying driver distraction and fatigue behaviors. It consists of multi-modal signals, including frontal images, ECG signals, and vehicle signals. To emphasize the practicality and authenticity of VTD, a platform was developed that integrates driving simulation and multi-modal signal acquisition functions to conduct experiments. Finally, the dataset was constructed through data processing and analysis to extract features related to driver state and behavior perception.


%

\subsection{Overall Framework} 

Driving data from 15 subjects in a fatigued condition and takeover experiment data from 17 distracted participants are included in the VTD dataset. All participants were fully informed about the research background and procedures, and they consented to participate by signing a written informed consent form. Detailed information about the subjects is provided in Table I. This paper will separately describe the specifics of the two data collection experiments and the data processing procedures.

\renewcommand{\arraystretch}{1.3} 
\begin{table}[h]
	\centering
	\begin{tabular}{c|cccccc}
		\Xhline{1.2pt} 
		\textbf{Experiment}  & \multicolumn{2}{c}{\textbf{Gender}}                                & \multicolumn{2}{|c}{\textbf{Age}}                          & \multicolumn{2}{|c}{\textbf{Driving Experience}}           \\ \hline
		            & male          & \multicolumn{1}{|c}{female} & \multicolumn{1}{|c}{mean} & \multicolumn{1}{|c}{SD} & \multicolumn{1}{|c}{mean} & \multicolumn{1}{|c}{SD} \\ \Xhline{0.75pt} 
		Fatigue     &    12           &         3                    &        32.2                   &        6.14                 &  3.73                         &        3.16                 \\ \hline
		Distraction & 14            & 3                           & 33.7                      & 6.45                    & 4.61                      & 3.65                    \\
          \Xhline{1.2pt} 
	\end{tabular}
	\caption{Basic Information of Participants}
	\label{tab:my-table}
\end{table}

\begin{figure*}[htbp]
  \centering
  \includegraphics[width=1\textwidth]{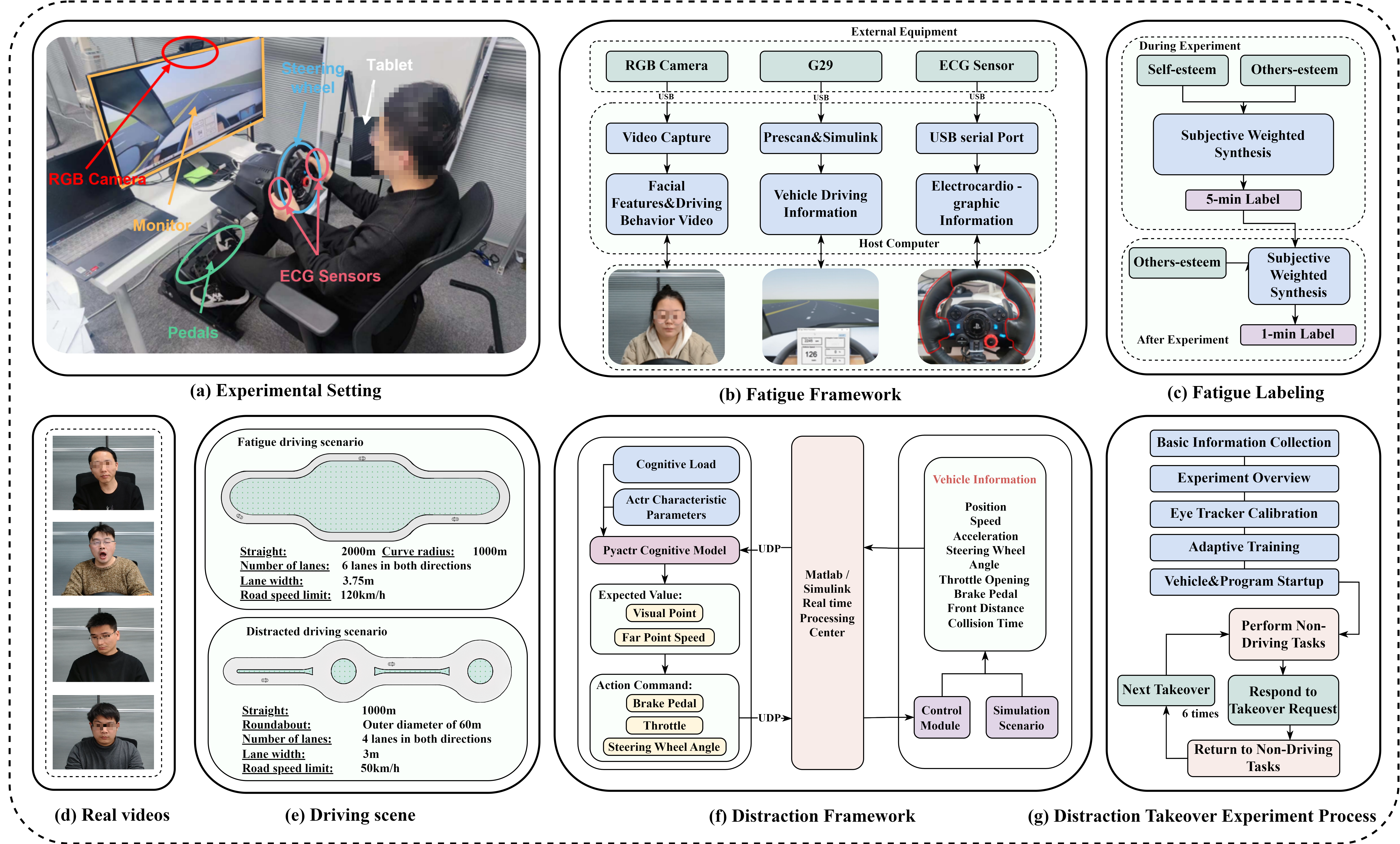}
  \caption{ VTD Data Collection Infrastructure}
  \label{fig:example}
\end{figure*}

Figure II summarizes the VTD experimental process and data collection infrastructure, as illustrated. (a) Physical diagram of the data acquisition platform, including the monitor, G29 steering wheel and pedal set, RGB camera, and ECG equipment. (b) Framework for fatigue driving experiments. (c) Flowchart for generating fatigue data labels. Data is labeled based on subjective and objective evaluations to assess driver behavior. (d) Examples of facial videos of drivers in distracted and fatigued states during driving simulation. (e) Experimental setup for fatigue and distraction driving models\footnote{To trigger distraction and fatigue, the scene is sparsely populated with vehicles, providing a wide driving view.}. (f) Framework for distracted driving experiments. (g) Process for distracted driving takeover experiments.


\subsection{Fatigue Driving Experiment}

The Fatigue dataset includes a processed time-series dataset and a raw simulation scenario video dataset. The time series dataset comprises eleven dimensions and is divided into a training set, a validation set, and a testing set in a ratio of 4:1:1.The total number of valid samples is 480, with a time slice of 60 seconds and a frame number of 1800. The dataset labels were modified based on KSS and SSS fatigue scales, and subjective evaluations were completed.

To construct a driver fatigue detection dataset, we collected information from participants and assigned pre-fatigue status according to age, sleep duration, and napping habits. Table II shows the pre-fatigue status of the participants. Participants were required to complete the pre-fatigue accumulation according to the assigned status. During the testing stage, the functionality of the platform connection and the data collection program was verified. Participants were instructed to complete the experiment preparations and driving adaptation under the operation instructions. They then performed the experiment wearing eye movement equipment and holding the electrode area of the steering wheel. They entered the appointed conditions and drove for 40 minutes continuously, during which they should keep their hands on the electrodes on the two sides. When the staff issued the prompt "report the current status" every five minutes, the participants should complete their self-evaluations while the staff completed their assessments on the participants, combining the observation results and reported results.

We generalized the different types of data collected as video data, vehicle data, and ECG data. The data were processed in different ways according to their characteristics.

For the driver frontal behavior information, the Mediapipe Facemesh model\cite{kartynnik2019real} was adopted for face landmark estimation,extracting 478 key coordinates of the driver’s face\footnote{Facial coordinate information can refer to: \url{https://github.com/google-ai-edge/mediapipe/blob/master/mediapipe/modules/face_geometry/data/canonical_face_model_uv_visualization.png}}
and characterized their mouths and eyesby Mouth Aspect Ratio (MAR) and Eye Aspect Ratio (EAR). The MAR and EAR values of each image frame are concatenated to form a time series. As indicated by the key points in the figure, the MAR and EAR are calculated as follows:
\begin{multline} 
\hspace*{-10pt}
MAR=\frac{\parallel\ P_{82}- P_{87} \parallel_{2} + \parallel\ P_{312}- P_{317} \parallel_{2}}{4\parallel\ P_{78}- P_{308} \parallel_{2}}
\end{multline}
\begin{multline} 
\hspace*{-10pt}
EAR=\frac{\parallel\ P_{387}- P_{373} \parallel_{2} + \parallel\ P_{385}- P_{380} \parallel_{2}}{4\parallel\ P_{263}- P_{362} \parallel_{2}} \\
+\frac{\parallel\ P_{158}- P_{153} \parallel_{2} + \parallel\ P_{160}- P_{144} \parallel_{2}}{4\parallel\ P_{133}- P_{33} \parallel_{2}}
\end{multline}

\renewcommand{\arraystretch}{1.35} 

\begin{table}[h]
\centering
\begin{tabular}{m{0.1\textwidth}|m{0.35\textwidth}}
\Xhline{1.2pt} 
\textbf{State} & \textbf{Requirements} \\
\Xhline{0.75pt} 
\textbf{A1} & Regular sleep: Ensure normal adequate sleep the night before the experiment. The experiment can be conducted in both the morning and afternoon. If in the afternoon, the nap time should be more than 0.5 hours. \\
\hline
\textbf{A2} & Nap deprivation: Ensure normal adequate sleep the night before the experiment. Conduct nap deprivation and the experiment only in the afternoon. \\
\hline
\textbf{A3} & Partial night sleep deprivation: Deprived of 40\%-60\% sleep the night before the experiment. No additional sleep on the day of the experiment. The experiment could be conducted in the morning and afternoon. \\
\hline
\textbf{A4} & Total night sleep deprivation: Deprived of over 80\% sleep the night before the experiment. No additional sleep on the day of the experiment. The experiment could be conducted in the morning and afternoon. \\
\Xhline{1.2pt} 
\end{tabular}
\caption{The Pre-Fatigue States of the Participants}
\end{table}

For driver head pose capture, we used Euler angles to characterize the rotational pose of the head. The Euler angle is derived by utilizing the facial key point information obtained from Facemesh and solving the rotation matrix by the PnP (Perspective-n-Point) algorithm. Note that the camera needs to be calibrated because its parameters affect the mapping relationship of the spatial points in the plane. This study employs a planar tessellated grid, calibrated using the Zhang Zhengyou calibration method to obtain the camera’s internal parameters and distortion coefficients. 

During driving, the steering wheel installed with flexible electrodes can be connected to the host computer via USB, and the driver’s ECG signals are acquired in real time by holding the steering wheel, with a sampling frequency of 250 Hz. This paper focuses on processing and feature extraction of ECG signals,primarily concentrating on the R-wave and utilizing the R-R intervals to compute the characteristics of heart rate and heart rate variability. For heart rate variability, the standard deviation of the R-R interval (SDNN) and the root mean square of the difference of the R0R interval (RMSSD) are used as metrics. To address baseline drift, IF interference, and EMG noise that occurr during the acquisition process, a Butterworth bandpass filter is applied.

For the vehicle data, we obtained them during real-time driving through the control signal of G29 with a sampling frequency of 100Hz, including the steering wheel angle, gas pedal signal, brake pedal signal, vehicle speed, and vehicle traverse angle speed.


\subsection{Distracted Driving and Takeover Experiment}

Regarding the driving takeover dataset, each participant is asked to perform three takeovers during visual and auditory subtasks in the investigation. Initially, the vehicle is in the autonomous driving phase while participants perform non-driving tasks on a tablet computer. When the system prompts a message to take over, drivers operate the vehicle while staff record relevant data and address unexpected situations.

This paper establishes 34 groups of visual and auditory subtasks for autonomous vehicles under three conditions (straight path, roundabout cut-in, and roundabout obstacle avoidance) and conducts 102 takeover experiments. After data screening and criteria extraction through nodes, takeover segments are extracted and divided according to the time nodes marking the start and end of the entire process, as well as those of the takeover.

Regarding takeover time, it is divided into takeover reaction time and takeover execution time. Takeover reaction time refers to the duration between the system’s takeover request and the driver’s return to the driving task (both hands back on the steering wheel), while takeover execution time is the sum of the duration during which the steering wheel angle$\geq2^\circ$ and the pedal was pressed $\geq 10\%$. 

VTD’s takeover and distraction data can also be used to calculate a driver’s load rate in human-vehicle co-pilot tasks through cognitive architecture models like QN-ACTR. Based on the load rate, driver’s fatigue level can be assessed, and by combining the vehicle’s displacement information, a safer and more reasonable human-vehicle driving right switching strategy can be designed.VTD also includes unscreened raw time series, raw videos, and tactile data so that users can filter and combine data based on research needs and goals.

\subsection{VTD Experiment Setup Innovation}

\subsubsection{Multi-channel and Multi-angle Videos}

Owing to the lack of public driver behavior datasets, most datasets are single-mode (RGB). For safety reasons, only simple visual signals can be collected in actual driving processes. These visual features usually depend on cameras and sensors directed toward the driver to obtain input data. The large-scale multi-view multi-modal database we constructed, VTD, can fill the gap for single visual signals. The accuracy of features extracted from facial detection, head pose estimation, and eye status analysis can be enhanced using multi-view information like driving view, eye movement, and facial view \cite{yuen2016looking}. Moreover, eye trackers’ high sampling rate, high precision, and low noise are advantageous compared to visual feature detection. Other modal features can be tuned to enhance the overall recognition rate using multi-view feature extraction and fusion.

Eye detection and eye status analysis are crucial to driver distraction and fatigue detection. Head rotation and eye closure rate can be calculated by applying PERCLOS to measure a driver’s fatigue level and PERLOOK \cite{jo2011vision} to measure ametropia duration. Due to limitations in resolution, camera-eye distance, and lighting conditions, it is not easy to calculate and distinguish the accuracy of data results from current mainstream datasets. However, VTD adopted Tobii Glasses3 to obtain omnidirectional eye movement tracking data from various angles, thus achieving the capture of high-precision eye movement data in an extensive range.

In current mainstream research methods, behavior analysis and fatigue detection have also been conducted by fully using the drivers’ diverse characteristics. These include detecting physiological signals, such as using EOG(Electrooculography) and ECG\cite{jo2011vision,wang2019modeling} or combining driving measurements (Steering wheel angle, steering speed, accelerator pedal angle, etc.) \cite{li2019fatigue,lim2016driver}. In all of the above scenarios, VTD is adaptable.

\subsubsection{Tactile sensing device for driver’s ECG}

To minimize the impact of the ECG devices on the driver, these devices are fixed on the G29 steering wheel. Signals are acquired through two flexible electrodes and transmitted to the collection program via USB, where they are saved as real-time texts. In contrast to signal acquisition from the participants’ left and right earlobes, participants only need to hold the electrode area of the steering wheel to perform real-time heart rate detection. This approach significantly reduces the chance of distraction and mitigates the devices' impact on the experiment.

\subsection{Data Processing Method Innovations}

To better align the dataset with the training model, normalization preprocessing is performed on the time-series data, which is then categorized according to research directions. Regarding driving fatigue detection, the VTD dataset contains 11-dimensional time series information and includes data series of the drivers’ frontal image, ECG signals, and the vehicle’s motion state. Fatigue levels are then graded by subjective evaluations combined with self-assessment and other’s assessment, thus realizing data calibration of Human-in-the-loop. Subsequently, dimension reduction and screening are performed on the above data to ensure a strong correlation between the data and the driver behaviors.

\renewcommand{\arraystretch}{1.35} 
\begin{table*}[h]
    \centering
    \resizebox{\textwidth}{!}{
    \begin{tabular*}{\textwidth}{@{\extracolsep{\fill}}c|c|c|c|c|c@{}}
        \Xhline{1.2pt}
        \textbf{Time series} & \textbf{Signal} & \textbf{Clue} & \textbf{F} & \textbf{P} & \textbf{S} \\
        \Xhline{0.75pt}
        \multirow{2}{*}{EAR} & \multirow{6}{*}{Driver Frontal Image Signal} & PERCLOS & 10.2095 & 4.3141$\times10^{-6}$ & ++++ \\
                             & & Blinking Rate & 4.2819 & 5.5569$\times10^{-3}$ & ++++ \\
        \cline{1-1} \cline{3-6}
        MAR                  & & MAR(SD) & 4.0222 & 8.1897$\times10^{-3}$ & ++++ \\
        \cline{1-1} \cline{3-6}
        Head Tilt            & & Head Tilt(SD) & 14.1214 & 2.0341$\times10^{-8}$ & ++++ \\
        \cline{1-1} \cline{3-6}
        Head Yaw             & & - & - & - & - \\
        \cline{1-1} \cline{3-6}
        Head Roll            & & - & - & - & - \\
        \hline
        \multirow{2}{*}{R-R} & \multirow{2}{*}{ECG} & SDNN & 12.3479 & 1.7280$\times10^{-7}$ & ++++ \\
                             & & RMSSD & 14.1791 & 7.8400$\times10^{-9}$ & ++++ \\
        \hline
        Steering Wheel Angle & \multirow{4}{*}{Vehicle Signals} & Steering Angle(SD) & 6.4363 & 5.5569$\times10^{-2}$ & +++ \\
        \cline{1-1} \cline{3-6}
        Pedal                & & Pedal(SD) & 3.5743 & 1.4349$\times10^{-2}$ & ++++ \\
        \cline{1-1} \cline{3-6}
        Vehicle Speed        & & Speed(SD) & 7.0730 & 1.2934$\times10^{-4}$ & ++++ \\
        \cline{1-1} \cline{3-6}
        Transverse Angular Velocity & & Transverse Angular Velocity(SD) & 5.7549 & 1.6380$\times10^{-4}$ & ++++ \\
        \Xhline{1.2pt}
    \end{tabular*}}
    \caption{Analysis of 10-Dimensional Time Series Information and F, P, and Significance Levels in Fatigue Data}
    \label{tab:my-table}
\end{table*}

\renewcommand{\arraystretch}{1.35} 

\begin{table*}[h]
    \centering
    \begin{tabularx}{\textwidth}{c|c|c|c|c}
        \Xhline{1.2pt} 
        \textbf{Dataset}   & \textbf{Features}      & \textbf{People} & \textbf{Quantity} & \textbf{Environment} \\  \Xhline{0.75pt} 
        \multicolumn{5}{c}{\hspace{3cm}\textbf{PUBLIC DISTRACTION DATASETS}}             \\ 
        \Xhline{0.75pt} 
        3MDAD\cite{jegham2020novel} & Comprehensive & 50   & 507 min videos, 20-34 sec each & Act + Real  \\ \hline
        DMD\cite{ortega2020dmd} & Comprehensive & 37   & 41h RGBD+IR videos & Real + Lab  \\ \hline
        VIVA\cite{das2015performance} & Hands & 8   & 2000+ images & Act  \\ \hline
        DriveAHead\cite{schwarz2017driveahead} & Heads & 20     & 100 million Depth \& IR images & Real  \\ \hline
        DAD\cite{kopuklu2021driver} & Behavior & 31     & 783 min videos & Act  \\ \hline
        MDAD\cite{jegham2019mdad} & Driver action & 50     & 2x2x800 video sequences & Real + Lab  \\ \hline
        \textbf{Ours VTD} & \textbf{Comprehensive} & \textbf{17}     & \textbf{6 types of scenarios, 102 takeover experiments,630 min videos} & \textbf{Real + Lab}  \\ 
        \Xhline{0.75pt} 
        \multicolumn{5}{c}{ \hspace{3cm}\textbf{DMS PUBLIC FATIGUE DATASETS}}             \\ 
        \Xhline{0.75pt} 
        YawDD\cite{abtahi2014yawdd} & Yawning & 107   & 342 videos, 15-40 sec each & Act + Real  \\ \hline
        ZJU\cite{pan2007eyeblink} & Eye Blinking & 20   & 80 videos & Act + Lab \\ \hline
        NTHU\cite{weng2017driver} & Drowsiness &  36  & 360 videos, 1 min each & Act + Simulated  \\ \hline
        RLDD\cite{ghoddoosian2019realistic} & Drowsiness & 12     & 180 videos, 10 min each & Real + Lab  \\ \hline
        NTHU-DDD\cite{weng2017driver} & Comprehensive & 36     & RGB + TXT & Act  \\ \hline
        DMD\cite{ortega2020dmd} & Comprehensive & 37   & 41h RGBD+IR videos & Real + Lab  \\ \hline
        CMU-PIE\cite{diaz2016reduced} & Head Pose & 72     & 1503 images & Real + Lab \\ \hline
        \textbf{Ours VTD} & \textbf{Comprehensive} & \textbf{15}     & \textbf{600 min videos,10-Dimensional Time-Series Signals} & \textbf{Real + Lab}  \\ 
        \Xhline{1.2pt} 
    \end{tabularx}
    \caption{Comprehensive Summary and Comparison of Public Fatigue and Distraction Datasets with VTD}
    \label{tab:my-table}
\end{table*}

Table III\footnote{"++++" represents very significant differences ($\alpha < 0.01$); "+++" represents significant differences ($0.01 \leq \alpha < 0.05$).} presents the 10 dimensional sequence information and analysis results of VTD fatigue data. The time series in some dimensions are chosen and investigated using One-way ANOVA (Analysis of Variance) to determine whether time series features are salient under different fatigue levels. From the ten features analyzed, VTD’s fatigue data and driver fatigue are strongly correlated\footnote{The F-value in Table III needs to be averaged over the serial information of each dimension and the total data, after which the specific value is obtained by calculating the ratio of the between-groups variance (MSA) to the within-groups variance (MSE). We assume that it satisfies the distribution $F(k-1,n-k)$, and obtain the probability $P$ based on the $F$ distribution, setting the significance level $\alpha=0.05$  as the benchmark. When $p<\alpha$ , it is considered that there is a significant difference between different serial data.}. All are valid inputs for the fatigue classification model.

This paper examines the differences in various takeover scenarios and subtasks under conditions of distraction and takeover. The findings indicate significant differences in collision avoidance conditions between straight roads and roundabouts ($P=5.536\times 10^{-10}<0.05$,$P=2.879\times 10^{-2}<0.05$); there are also significant differences under visual and auditory subtasks ($P=9.120\times 10^{-4}<0.05$,$P=6.060\times 10^{-3}<0.05$). These results suggest that different subtasks and takeover scenarios impact takeover reaction time.

\section{PROPERTIES}

\subsection{Video Data Characteristics and Applications}

VTD contains various complex combinations of visual indications and different fatigue levels and distractions in takeover tasks. The experiment includes 10-hour fatigue driving data from 15 participants and 102 takeover multidimensional experiment data from 17 participants (including recorded video data).Table IV lists a comparison of attributes between VTD and existing datasets. Compared to other publicly available data, VTD offers multimodal, multi-view, diverse, and fine-grained data that is controllable and quantifiable. This provides rich data support for research on human-machine collaborative driving systems and driver safety monitoring systems.


Another key highlight of VTD is the construction of a long-sequence multimodal natural data based on visual-tactile data fusion. We designed extended time-series segments integrating multiple modalities, including RGB facial video, vehicle motion data,  tactile ECG data and images captured by the eye trackers in driving scenarios. This comprehensive approach enhances data diversity and granularity, providing critical insights into driver fatigue and distraction.These video data can be utilized in studies including driver fatigue detection, distraction monitoring, and human-vehicle driving control transitions.


\subsection{Properties and Functions of Tactile Data}

VTD provides piezoelectric tactile ECG data and steering wheel data. In the 40-minute experiment, we gathered drivers’ ECG and PPG data using their tactile feedback to the flexible electrode during real-time driving. At the same time, 7-dimensional data of the steering wheel, the vehicle direction, the brake, the accelerator, the gear position, and the turning angle were collected. The driver’s feedback and steering wheel data formed cross-validation.

Traditional visual driving behavior detection methods are limited by lighting conditions and the vehicle’s location. Consequently, they are unable to satisfy continuous, high-quality visual signal collection. Additionally, visual identity systems are also faced with problems including but not limited to computing power issues and communication delays. We can partly solve the above issues with the wearable piezoelectric tactile device without creating constraints or disturbance caused by traditional wearable devices. However, difficulties still exist, for example, too many environmental interference sources and insufficient robustness for dynamic changes in signals and environments\cite{boon2015mobile,lee2016wearable}. A breakthrough that future driver behavior perception technology should anticipate is a combination of vision and tactile sensations that can balance the driver’s state, the accuracy of behavior recognition, and application adaptability.

\subsection{Properties and Functions of Visual-tactile Combinations}

Combining vision and tactile sensation can compensate for the robustness of visual perception by incorporating the visual modality’s sensitivity to position and movement and the tactile modality’s rapidity. It can reduce the system delay under the risk of data overload and form mutual complementary effects under vehicle tracking and collision avoidance control \cite{yi2022survey}. Visual-tactile fusion requires temporal embedding when combined with multi-dimensional time-series data. Positional embeddings are also added to non-linear transformed time series to leverage the sequential correlations based on time steps. While using Transformer to classify time series and make predictions, positional embeddings can be employed to solve the scene adaptation issue of position data and time series in Transformer. These positional embedding vectors, along with multi-dimensional time series, can be injected into the model as additional input.

\section{CONCLUSIONS}

This paper presents a method for constructing a long-sequence multimodal natural dataset based on visual-tactile data fusion. The aim is to provide data support for quantifying and validating drivers’ fatigue and distraction detection across identical driving scenarios, as well as for cross-modal perception algorithms related to driver behaviors, such as driving takeover monitoring. To meet various research demands, the VTD dataset includes data on fatigue driving and the drivers’ visual and tactile behaviors during human-vehicle driving control transitions. This work aims to establish a standardized platform for benchmark testing, thereby advancing the development of driver behavior perception and enhancing research on driving safety.


\section*{ACKNOWLEDGMENT}

The authors would like to thank all the participants for their great support in the data acquisitions, and especially thank the Kaiwu Laboratory of Shanghai Research Institute for Intelligent Autonomous Systems at Tongji University for providing experimental conditions.

\bibliographystyle{IEEEtran} 
\bibliography{root}          

\end{document}